\documentclass[11pt,a4paper]{article}
\usepackage[hyperref]{emnlp-ijcnlp-2019}
\usepackage{times}
\usepackage{latexsym}
\usepackage{listings}
\usepackage{mathtools}
\usepackage{multirow}

\usepackage{url}

\aclfinalcopy

\title{ASU at TextGraphs 2019 Shared Task:  Explanation ReGeneration using Language Models and Iterative Re-Ranking }

\author{Pratyay Banerjee
\\ School of Computing, Informatics, and Decision Systems Engineering, Arizona State University
\\ \texttt{pbanerj6}@asu.edu
}

\date{}

\begin{document}
\maketitle
\begin{abstract}
 In this work we describe the system from Natural Language Processing group at Arizona State University for the TextGraphs 2019 Shared
Task. The task focuses on Explanation Regeneration, an intermediate step towards general multi-hop inference on large graphs. Our approach consists of modeling the explanation regeneration task as a \textit{learning to rank} problem, for which we use state-of-the-art language models and explore dataset preparation techniques. We utilize an iterative re-ranking based approach to further improve the rankings. Our system secured 2nd rank in the task
with a mean average precision (MAP) of 41.3\% on the test set.
\end{abstract}

\section{Introduction}
Question Answering in natural language often requires deeper linguistic understanding and reasoning over multiple sentences. For complex questions, it is very unlikely to build or have a knowledge corpora that contains a single sentence answer to all the questions from which a model can simply cherrypick. The knowledge required to answer a question may be spread over multiple passages \cite{squad2,race}. Such complex reasoning requires systems to perform multi-hop inference, where they need to combine more than one piece of information \cite{welbl2018constructing}.

\begin{table}[!ht]
\begin{center}
\begin{tabular}{|p{7cm}|}
\hline \textbf{Question:} \textit{Which of the following is an example of an organism taking in nutrients?}\\
(A) a dog burying a bone (B) \textbf{a girl eating an apple} (C) an insect crawling on a leaf (D)  a boy planting tomatoes \\  
\hline
\textbf{Gold Explanation Facts}:\\
A girl means a human girl. : Grounding\\
Humans are living organisms. : Grounding \\
Eating is when an organism takes in nutrients in the form of food. : Central \\
Fruits are kinds of foods. : Grounding \\
An apple is a kind of fruit. : Grounding \\
\hline
\textbf{Irrelevant Explanation Facts}:\\
Some flowers become fruits. \\
Fruit contains seeds. \\
Organisms; living things live in their habitat; their home \\
Consumers eat other organisms \\
\hline
\end{tabular}
\end{center}
\caption{\label{tab1} An example of Explanation Regeneration }
\end{table}

In this shared task of \textit{Explanation ReGeneration}, the systems need to perform multi-hop inference and rank a set of explanatory facts for a given elementary science question and correct answer pair. An example is shown in Table \ref{tab1}. The task provides a new corpora of close to 5000 explanations, and a set of gold explanations for each question and correct answer pair \cite{jansen2018worldtree}. The example highlights an instance for this task, where systems need to perform multi-hop inference to combine diverse information and identify relevant explanation sentences needed to answer the specific question.

\textit{Explanation ReGeneration} is a challenging task. This is due to the presence of other irrelevant sentences in the corpora with respect to the given question, which have a good lexical and semantic overlap \cite{jansen2018multi}. Ideally, the explanations need to be in order, but for the sake of simplicity, ordering of the explanations are ignored.

In the dataset, to measure the performance of the system over different types of explanations, the explanations are further categorized into classes. These classes differ in the importance of the explanation in explaining the correct answer. These categories are \textit{Central}, \textit{Grounding} and \textit{Lexical Glue}. Central facts are often core scientific facts relevant to answering the question. Grounding are facts which connect to other core scientific facts, present in the explanation. Examples of Central and Grounding facts are present in Table \ref{tab:my-table}. Lexical glue facts express synonymy or definitional relationships. An example of Lexical glue facts : ``glowing means producing light".

This paper describes a system developed by the Natural Language Processing group of Arizona State University.  We approach the task of explanation regeneration as a \textit{learning to rank} \cite{burges2005learning} problem. The system utilizes state of the art neural language models \cite{bert,xlnet}, finetunes them on the knowledge corpora and trains them to perform the task of ranking using customized dataset preparation techniques. We further improve on the ranking using an iterative re-ranking algorithm. 

We make the following contributions in the paper: (a) We evaluate different ways for dataset preparation to use neural language models for the task of explanation generation; (b) We evaluate different language models for ranking and analyse their performance on the task; (c) We show how to use iterative re-ranking algorithm to further improve performance; (d) We also provide a detailed analysis of the dataset.

In the following sections we first give an overview of our system. We describe the individual components of the system in Sections \ref{sec:dprep},\ref{sec:iter}. We evaluate each component on the provided validation set and show the performance on the hidden test set in Section \ref{sec:exps}. We conclude with a detailed error analysis and evaluating our model with relevant metrics in Section \ref{sec:error},\ref{sec:disc},\ref{sec:conc}.

\section{Approach}
In recent years, several language models \cite{bert,xlnet,elmo} have shown considerable linguistic understanding and perform well in tasks requiring multi-hop reasoning such as question answering \cite{squad2,obqa,multirc,race}, and document ranking tasks \cite{clueweb09}.

For the task of \textit{Explanation ReGeneration} we chose BERT \cite{bert} and XLNET \cite{xlnet}, two state-of-the-art neural language models and explore their effectiveness in capturing long inference chains and performing multi-hop inference. BERT and XLNET are pretrained using Masked Language Modelling (MLM)\label{mlm} and Probabilistic Masked Language Modelling (PMLM) \label{pmlm} respectively. These pretraining tasks enables BERT and XLNET to understand the dependencies between masked and unmasked words. This is needed to capture relevant concepts, words and entities linking between central, grounding and lexical glue facts. We finetune these language models on the knowledge corpora of 5000 explanations using their respective language modelling tasks. 

To rank the explanations, we learn the relevance of each explanation for a given question and correct answer pair. We evaluate multiple dataset preparation techniques for finetuning the language models. We also evaluate different relevance learner models by attaching different kinds of classification and regression heads over the language models. From the relevance learner, we obtain the \textit{relevance} scores and an initial ranking of the explanations. We perform further re-ranking using a custom re-ranking algorithm similar to in \citeauthor{banerjee2019careful}.

\section{Dataset Preparation and Relevance Learner}
\label{sec:dprep}
We prepare multiple datasets for the following tasks. The preparation techniques are described in the following sub-sections.

\subsection{Language Modelling}
The language models are initially finetuned on the Explanation Knowledge corpora using MLM and PMLM respectively. The dataset for this task is prepared using scripts from \href{https://github.com/huggingface/pytorch-transformers}{pytorch-transformer} package. We prepare both MLM\label{mlm} and PMLM \label{pmlm} datasets and finetune the respective language models. We follow the steps as mentioned by \citeauthor{bert} and \citeauthor{xlnet} for generating the language model datasets.

\subsection{Relevance Learning using Classification head}
We model the relevance learning task as a two-class classification task with class 0 representing irrelevance and class 1 representing relevance. Here by relevance, we mean the fact is part of the explanation. Finally, we take the probability scores of class 1, and use them as relevance scores. Formally,
\begin{equation}
    Rel(E_j,Q_i,A_i) = P(E_j \in G|Q_i,A_i)
\end{equation}
where $E_j$ is the $j$th explanation, $Q,A$ are the $i$th question and correct answer pair and $G$ is the set of gold explanation facts.  

The dataset provides the set of gold facts, but does not provide the set of irrelevant facts. To create the irrelevant set, for each gold fact we retrieve top $k$ similar facts present in the explanation corpora, but not present in the gold fact set. This is done to make the model learn relevance to the context of passage and correct answer, and not focus on similar looking sentences. We compute the similarity between sentences using \textit{cosine similarity} between sentence vectors \cite{spacy2}. We repeat the gold explanation $k$ times to maintain the balance between the classes. 

We prepare another dataset where we provide a context passage. This passage comprises of already selected $n$ facts, and the rest $|G| - n$ gold facts are labelled as relevant class 1. We find irrelevant facts for the $|G| - n$ gold facts using the same process as above. In this case, we learn the following probability:
\begin{equation}
    Rel(E_j,C_n,Q_i,A_i) = P(E_j \in G|C_n,Q_i,A_i)
\end{equation}
where $C_n$ represents the $n$ already selected facts, $1 \leq n \leq 16$, as there are a maximum of 16 and minimum of 1 gold explanation facts. This context is given only during the \textit{training phase}, while during the validation and testing, we only provide the question and correct answer pair along with explanation $E_j$. Moreover, we ensure that the dataset is balanced between two classes. To make the model learn longer dependencies, we train using a context. This classification task optimizes classification accuracy metric.

\subsection{Relevance Learning using Regression head}
The datasets for the regression tasks are similar to the datasets of classification head. Instead of two class classification, we provide target scores of 6, 5, 4, 0 for Central, Grounding, Lexical Glue and Irrelevant facts respectively. The above scoring scheme was decided to give central and grounding facts higher precedence, as they are core for a proper explanation. All target scores were ensured to be balanced. As described in the above section, we prepare two datasets, one with and another without context explanation sentences. The regression task optimizes mean-squared-error scores.

\section{Iterative Re-Ranking}
\label{sec:iter}
We sort the Relevance scores from the Relevance Learner models and perform an initial ranking of the explanation sentences. We feed this initial ranked explanation facts to our iterative re-ranking algorithm, which is defined as follows.
    
Let $N$ be the depth of re-ranking from the top, i.e, we run re-ranking for $N$ rounds. Let $E_0$ be the top explanation fact in the initial ranking, $E_i$ be the last selected explanatory fact and $E_j$ ($i<j \leq N+i$) is the current candidate explanation fact for a given question $Q$ and correct answer $A$. We compute a \textit{weighted relevance score} using the top $i$ ($i<=N$) selected facts as:
\begin{equation}
    W_{score}(E_j,E_i) = \frac{\sum_{k=0}^i{Rel(E_k)*Sim(E_j,E_k)}}{\sum_{k=0}^i{Rel(E_k)}}
\end{equation}

We sort ranking scores of the candidate explanation facts and choose the top explanation fact for the $i+1$ th round, where the ranking score is given by :
\begin{equation}
\begin{split}
        & score(E_j,E_i,Q,A) \\
        &= W_{score}(E_j,E_i) * Sim(E_j,Q:A)
\end{split}
\end{equation}
Here $Rel$ is the relevance score from the Relevance Learner models and $Sim$ is the cosine similarity of the explanation sentence vectors from Spacy \cite{spacy2}. For the facts whose initial rank is greater than depth $N$, we keep the initial ranking as is. The iterative re-ranking algorithm is designed to exploit the overlapping nature of the explanations. The above score gives importance to the initial relevance score (facts already ranked by relevance), the vector similarity of the candidate explanation and both the selected explanation sentences and question/correct answer pair.

\section{Experiments and Test Results}
\label{sec:exps}
The training dataset for the task contains 1191 questions, each having 4 answers. The gold explanations set has a minimum size of 1 and maximum size of 16. The validation dataset had 265 questions and the hidden test set has 1248 questions. The explanation knowledge corpora has around 5000 explanation sentences. The two relevance learner training dataset has size of 99687 without context and 656250 with context. Several combinations of context are generated using the gold selected explanation facts, leading to such a large training corpus. We evaluate both BERT Large and XLNET large, using both the tasks and the two different datasets. In Table \ref{tab:my-table} are the results of our evaluation on the validation set. All the metrics are Mean-Average-Precision unless mentioned otherwise. All the metrics are on the validation set.

\begin{table}[h]
\centering
\begin{tabular}{|l|l|l|}
\hline
Task v/s Model                  & BERT    & XLNET   \\ \hline
Classification              & 0.3638 & 0.3254 \\ \hline
Classification with Context & \textbf{0.3891 }& 0.3473 \\ \hline
Regression                  & 0.3288 & 0.3164 \\ \hline
Regression with Context     & 0.3466 & 0.3273 \\ \hline
\end{tabular}
\caption{Comparison of the Relevance Learners with different dataset preparation techniques without re-ranking}
\label{tab:my-table}
\end{table}

It can be observed in Table \ref{tab:my-table} that the two class classification head with context performs best and BERT Large outperforms XLNET Large for this particular task.  In Table \ref{tab:my-table2}, we compare the Relevance Learners before and after Iterative Re-ranking. It can be seen that Iterative Re-ranking improves the scores of both the Relevance Learners by around 2.5\%.

\begin{table}[]
\centering
\begin{tabular}{|l|l|l|}
\hline
\begin{tabular}[c]{@{}l@{}}$N$ v/s Model\end{tabular} & BERT   & XLNET  \\ \hline
1                                                                    & 0.3891 & 0.3473 \\ \hline
3                                                                    & 0.4000 & 0.3556 \\ \hline
5                                                                    & 0.4062 & 0.3661 \\ \hline
10                                                                   & 0.4181 & 0.3701 \\ \hline
15                                                                   & \textbf{0.4225} & \textit{0.3738} \\ \hline
20                                                                   & 0.4204 & 0.3721 \\ \hline
30                                                                   & 0.4191 & 0.3665 \\ \hline
\end{tabular}
\caption{Comparison of Relevance Learners with Iterative Re-ranking till depth $N$}
\label{tab:my-table2}
\end{table}

 Table \ref{tab:my-table3} compares the MAP for different explanation roles before and after iterative re-ranking. It can be seen that Iterative re-ranking improves MAP for Central and Grounding explanations but penalizes Lexical Glue. 

\begin{table}[h]
\centering
\begin{tabular}{|l|l|l|}
\hline
Explanation Roles & BERT   &  $N$=15 \\ \hline
CENTRAL                     & 0.3589 & 0.3912             \\ \hline
GROUNDING                   & 0.0631 & 0.0965             \\ \hline
LEXICAL GLUE                & 0.1721 & 0.1537             \\ \hline
BACKGROUND \footnotemark                 & 0.0253 & 0.0226             \\ \hline
NEG \footnotemark[\value{footnote}]                     & 0.0003 & 0.000586           \\ \hline
\end{tabular}
\caption{MAP for different Explanation Roles for BERT trained with classification head and context, before and after re-ranking till $N$=15 }
\label{tab:my-table3}
\end{table}

\footnotetext{Background and Neg roles were found in the gold explanation set but definition for them are not shared.}

\begin{figure}[h]
  \includegraphics[width=22em]{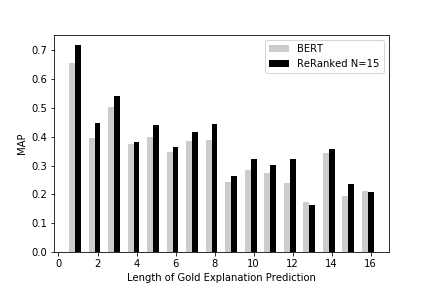}
  \caption{MAP v/s Length of the Gold Explanation}
  \label{fig:plt1}
\end{figure}

Figure \ref{fig:plt1} shows performance of the Relevance Learner and Iterative Re-ranking for questions with different length of gold explanations. It can be seen that the model performs well for explanations whose length are less than or equal to 5. Performance decreases with increasing length of gold explanations.

\begin{table}[]
\centering
\begin{tabular}{|l|l|l|}
\hline
Model             & Validation & Test   \\ \hline
BERT with Context & 0.3891     & 0.3983 \\ \hline
ReRanked N=15     &\textbf{ 0.4225}     & \textbf{0.4130} \\ \hline
Baseline SVM Rank & 0.28       & 0.2962 \\ \hline
\end{tabular}
\caption{Validation and Test MAP for the best Relevance Learner, Reranked and the provide baseline models }
\label{tab:my-table4}
\end{table}

Table \ref{tab:my-table4} shows the MAP scores for the best models on both the Validation and the hidden Test set. 

\section{Error Analysis}
\label{sec:error}

In the following sub-sections we analyse our system components, the performance of the final re-ranked Relevance Learner system  and the shared task dataset.

\begin{table*}[t]
\centering
\begin{tabular}{|l|l|}
\hline
\textbf{Gold Explanation}                   & \textbf{Predicted Explanation}                                 \\ \hline
heat means temperature increases   & adding heat means increasing temperature              \\ \hline
sunlight is a kind of solar energy & the sun is the source of solar energy called sunlight \\ \hline
look at means observe              & observe means see                                     \\ \hline
a kitten is a kind of young; baby cat              & a kitten is a young; baby cat                                     \\ \hline
\end{tabular}
\caption{Similar Explanations present in top 30}
\label{tab:similarexp}
\end{table*}

\begin{table*}[t]
\centering
\begin{tabular}{|c|c|}
\hline
\multicolumn{1}{|l|}{\textbf{Gold Explanation}}               & \multicolumn{1}{l|}{\textbf{Predicted Explanation}}        \\ \hline
\multirow{2}{*}{an animal is a kind of living thing} & an animal is a kind of organism                    \\ \cline{2-2} 
                                                     & an organism is a living thing                      \\ \hline
\multirow{2}{*}{a frog is a kind of aquatic animal}  & a frog is a kind of amphibian                      \\ \cline{2-2} 
                                                     & an amphibian is a kind of animal                   \\ \hline
a leaf is a part of a tree                           & \multirow{2}{*}{a leaf is a part of a,green plant} \\ \cline{1-1}
a tree is a kind of plant                            &                                                    \\ \hline
\end{tabular}
\caption{Single-hop and Multi-hop Errors in top 30}
\label{tab:multihop}
\end{table*}

\subsection{Model Analysis}
\begin{table}[t]
\centering
\begin{tabular}{|l|}
\hline
\textbf{Gold Explanation}    \\ \hline
to reduce means to decrease    \\ \hline
\textbf{Predicted Explanation} \\ \hline
to lower means to decrease     \\ \hline
less means a reduced amount    \\ \hline
\end{tabular}
\caption{Errors due to Sentence Vectors in top 30}
\label{tab:sentencevectors}
\end{table}


\begin{enumerate}
    \item XLNET performs poorly compared to BERT for this task. The difference arises due to two reasons, the way the datasets are prepared and the way the language models are finetuned. The dataset preparation techniques  BERT captures deeper chains and better ranks those explanation which have low direct lexical or semantic overlap with the question and correct answer.
    
    \item XLNET focuses on explanations mainly on the words whose word vectors are closely related to the question and answer, and performs poorly for explanations which are one or two hop away. The dataset with context, improves the performance for both, enabling capturing deeper chains to some extent.

    \item The way the datasets are prepared introduces bias against some explanation facts. For example, the Lexical Glue facts are of the type ``X means Y'' and the Grounding facts are of the type ``X is kind of Y''. Using sentence vectors for identifying similar but irrelevant explanations leads to a set of explanations being particularly tagged as irrelevant. These are ranked low even if they are relevant for the validation set. This leads to poor performance compared to Central facts.
    
    \item The Iterative Re-ranking algorithm improves the performance irrespective of the Relevance Learner model. The algorithm gives importance to the \textit{relevance score}, vector similarity with previously selected explanations and vector similarity with the question answer pair. This introduces a bias against Lexical Glue explanations, as they only have a word common with the entire question, answer and previously selected facts. The optimal depth of the re-ranking correlates with the maximum length of explanations.

    \item The model is able to rank Central explanations with a high precision. Central facts possess a significant overlap with the question and correct answer. The re-ranking algorithm improves the precision even further. From Figure \ref{fig:plt1}, it can be inferred, the explanations with length 1 only contain Central explanations. For explanations with length 2, the model precision drops considerably. This occurs because the model is poor in ranking Lexical Glue and Grounding explanations.
    
\begin{table}[t]
\centering
\begin{tabular}{|l|}
\hline
\textbf{Gold Explanation}                 \\ \hline
temperature rise means become warmer        \\ \hline
\textbf{Predicted Explanation}              \\ \hline
warm up means increase temperature          \\ \hline
warmer means greater; higher in temperature \\ \hline
\end{tabular}
\caption{Model unable to understand ordering in Lexical Glue in top 30}
\label{tab:lexicalorder}
\end{table}
    
    \item 
    \label{lexerror}
    For Lexical Glue and Grounding explanations, which have the form ``X means Y'' and ``X is kind of Y'', the model is not able to understand the order between X and Y required for the explanation, i.e, instead of ``X means Y'', it ranks ``Y means X'' higher. Table \ref{tab:lexicalorder} is one such instance. This contributes to the low MAP for these types of explanations.

    \item 
    \label{senterror}
    The use of sentence vectors for similarity introduces errors shown in Table \ref{tab:sentencevectors}, where the correct explanation contains ``reduce", but the predicted explanations which have similar words like ``lower'', ``less'' and ``reduced amount'', are ranked higher. 
    
    \item Out of the total 226 questions in the validation set, there were only 13 questions for which the system could not predict any of the gold explanation facts in the top 30. The facts predicted for these had a high word overlap, both symbolic and word-vector wise, but were not relevant to the explanation set.
    
\end{enumerate}

\subsection{Dataset Analysis}
\begin{enumerate}
    \item We further looked at the gold annotations and top 30 model predictions and identified few predictions having similar semantic meaning being present. For example in Table \ref{tab:similarexp}, the predicted explanations were present in top 30. This shows there can be alternate explanations other than the provided gold set.
    \item In Table \ref{tab:multihop} we can see the model makes both kinds of errors. For few gold explanations, it brings two alternate explanation facts and for some explanations it combines the facts and ranks a single explanation in the top 30. This also shows there can be several such combinations possible.
\end{enumerate}

\section{Discussion}
\label{sec:disc}
From the analysis, we can observe that multiple alternate explanations are possible. This is analogous to multiple paths being available for the explanation of a phenomenon. Our model precision should improve with availability of such alternate explanations. We recommend enriching the gold annotations with possible alternatives for future rounds of the task.

It is promising to see a language model based on stacked attention layers is able to perform multi-hop inference with a reasonable precision, without much feature engineering. The use of neural language models and sentence vector similarities bring errors, such as point \ref{lexerror} and \ref{senterror} in the Error Analysis section. We can introduce symbolic and graph based features to capture ordering \cite{witschel-2007-multi}. We can also compute graph feature-enriched sentence vectors using principles of textual and visual grounding \cite{cai2018comprehensive,guo2016jointly,yeh2018unsupervised,grover2016node2vec}. In our system design, we did not use the different explanation roles and the dependencies between them. Using such features the precision is likely to improve further. Our Iterative re-ranking algorithm shows it can improve the precision even more, given a reasonably precise Relevance Learner model. This is the first time this task has been organized and there is lot of scope for improvement in precision.

\section{Conclusion}
\label{sec:conc}
In this paper, we have presented a system that participated in the shared task on explanation regeneration and ranked second out of 4 participating teams. We designed a simple system using a neural language model as a relevance learner and an iterative re-ranking algorithm. We have also presented detailed error analysis of the system output, the possible enrichments in the gold annotations of the dataset and discussed possible directions for future work.


\bibliography{emnlp-ijcnlp-2019}
\bibliographystyle{acl_natbib}

\appendix

\section{Supplemental Material}
\label{sec:supplemental}

\end{document}